\documentclass[a4paper]{article}
\RequirePackage{amsmath}
\usepackage[utf8]{inputenc}
\usepackage[hidelinks]{hyperref}
\usepackage{amssymb}
\usepackage{mathabx}
\usepackage{natbib}
\usepackage{graphicx}
\usepackage{float}
\usepackage{authblk}
\usepackage[margin=1cm,font=small]{caption}
\title{Privacy-Preserving Generalized Linear Models using Distributed Block Coordinate Descent}

\author[1]{Erik-Jan van Kesteren\thanks{Correspondence about this article should be sent to 
		e.vankesteren1@uu.nl}}
\author[2]{Chang Sun}
\author[1]{Daniel L. Oberski}
\author[2]{Michel Dumontier\thanks{These authors share last authorship}}
\author[2]{Lianne Ippel$^\dagger$}
\affil[1]{Department of Methodology and Statistics, Utrecht University}
\affil[2]{Institute of Data Science, Maastricht University}

\date{\today}

\begin{document}
	
	\maketitle
	
	\begin{abstract}
		Combining data from varied sources has considerable potential for knowledge discovery: collaborating data parties can mine data in an expanded feature space, allowing them to explore a larger range of scientific questions. However, data sharing among different parties is highly restricted by legal conditions, ethical concerns, and / or data volume. Fueled by these concerns, the fields of cryptography and distributed learning have made great progress towards privacy-preserving and distributed data mining. However, practical implementations have been hampered by the limited scope or computational complexity of these methods. In this paper, we greatly extend the range of analyses available for vertically partitioned data, i.e., data collected by separate parties with different features on the same subjects. To this end, we present a novel approach for privacy-preserving generalized linear models, a fundamental and powerful framework underlying many prediction and classification procedures. We base our method on a distributed \textit{block coordinate descent} algorithm to obtain parameter estimates, and we develop an extension to compute accurate standard errors without additional communication cost. We critically evaluate the information transfer for semi-honest collaborators and show that our protocol is secure against data reconstruction. Through both simulated and real-world examples we illustrate the functionality of our proposed algorithm. Without leaking information, our method performs as well on vertically partitioned data as existing methods on combined data -- all within mere minutes of computation time. We conclude that our method is a viable approach for vertically partitioned data analysis with a wide range of real-world applications.
	\end{abstract}

	\section{Introduction}
	With technological developments in computational power and storage capacity, an increasing amount of data is collected and stored by a variety of data parties \citep{kaisler2013big}. Over the past decades, data mining has been successful in extracting information from such datasets, but it is especially powerful when various data sources are combined: collaborating data parties can mine data in a larger feature space, allowing them to discover knowledge beyond their individual potential. For example, in the medical domain, personal health conditions are significantly affected not only by genetic and biological factors, but also by individual behaviour and social circumstances \citep{world2008closing}; combining those sources has the potential to improve analytical models for health outcomes \citep{kasthurirathne2017assessing, ancker2018potential}. 
	
	However, there is a pertinent obstacle to unlocking the potential of combining datasets: integrating various sources may reveal private information about individual data subjects to the collaborating parties. Hence, data sharing is highly restricted by legal and ethical concerns. This highlights the need for privacy-preserving techniques which perform data mining tasks on multiple sources without explicitly sharing their full data \citep[e.g.,][]{du2004privacy, gambs2007privacy, karr2009privacy, gascon2017privacy}. In this paper, we develop a novel algorithm for performing generalized linear modeling (GLM) in a privacy-preserving way in such a partitioned data situation. GLM is a powerful statistical framework for prediction and classification and is at the basis of a wide range of analysis applications including linear, count, and logistic regression \citep{mccullagh1989generalized, dobson2008introduction}.
	
	This paper is organized as follows. In Section \ref{sec:rel}, related work is discussed to contextualize our contribution. In Section \ref{sec:method}, we introduce our proposed method for GLM on vertically partitioned data. Next, we describe in detail the privacy-preserving and information sharing characteristics of this protocol in Section \ref{sec:priv}, and we analyze how the information transfer affects the ability of the partner organisation to recover the collaborator's data. In Section \ref{sec:experiments}, we benchmark our implementation of the protocol against full-data analysis using Monte Carlo simulations and we illustrate the functionality of our implementation using three different real-life data sets from the UCI Machine Learning repository \citep{blake1998uci}. Finally, we discuss the strengths and limitations of our approach in Section \ref{sec:disc} and we provide suggestions for future research.
	
	All of the methods described here are implemented in \texttt{privreg}, an open-source software package for the \texttt{R} programming language \citep{RCoreTeam2018}. This implementation includes encryption for all communication across parties based on a pre-shared key, and includes a user-friendly interface based around an object-oriented architecture. The package is available for installation from \url{https://github.com/vankesteren/privreg}.
	

	\section{Related work}
	\label{sec:rel}
	In practice, there are two main types of data partitioning \citep{vaidya2005privacy}. Different data sources might collect the same features of different data subjects, e.g., different hospitals collect the same type of information from their own set of patients. This situation is referred to as \textit{horizontally partitioned} data. Alternatively, separate sources might collect different information from the same data subjects, e.g., medical features by the hospital may be combined with socioeconomic features from a government statistics department. This situation is referred to as \textit{vertically partitioned} data, which is the focus of the current paper. There is also a third scenario, where data are both vertically and horizontally partitioned, which may be referred to as \textit{hybrid partitioning}. 
	
	Our aim is to analyze data which is vertically partitioned without leaking raw data to the collaborating parties (\emph{Alice} and \emph{Bob}). In order to analyze such data, either the dataset may be combined but hidden from the collaborating parties, or the analytical procedure should prevent leaking of information. The former relies on the inclusion of an `uninterested' or trusted third party (TTP): Each party sends their raw data encrypted to the TTP, who then performs the required analyses on the combined data sets. Afterwards, the TTP returns the results to all data parties and the raw data of \emph{Alice} stays hidden to \emph{Bob}. However, this solution requires all parties to fully trust the TTP, which might not be possible in the face of restrictive legislation or sensitive data.  
	
	There is another class of methods which do not rely on a TTP, instead using cryptography to perform data mining tasks on vertically partitioned data. These methods focus on preventing information leakage by creating protocols which hide the raw data from the collaborator \citep[e.g., for the construction of decision trees, ][]{agrawal2000privacy}. In this class of methods \citet{du2001privacy} and \citet{du2004privacy} investigated various protocols for secure matrix computation for linear least squares regression and classification problems. Several other authors used and extended more general secure multiparty computation protocols \citep[e.g., the garbled circuit protocol;][]{yao1986generate} to perform regression on vertically partitioned data \citep{amirbekyan2007privacy, Slavkovic:2007:SLR:1335998.1336070, fang2013privacy, nikolaenko2013privacy, gascon2016secure, gascon2017privacy, bloom2019secure}. While their use of these general protocols yields certain privacy guarantees, their practical implementations and use are hindered by requiring semi-trusted third parties, intermediate data sharing, computational complexity, or a limitation to the linear regression situation. 
	
	Another line of research leverages the privacy-preserving properties of algorithms from \textit{federated} or \textit{distributed learning}, a field researching data mining on separated datasets \citep{li2019federated, dobriban2018distributed}. A canonical example is by \citet{Sanil:2004:PPR:1014052.1014139}, who developed a method to compute linear regression coefficients iteratively based on an algorithm by \citet{powell1964efficient}. Other authors leverage specific distributed learning algorithms to implement statistical learning for vertically partitioned data \citep{vaidya2002privacy, vaidya2003privacy, vaidya2005privacy, vaidya2008privacy}. Our method is closely related to this branch of research. Unlike existing regression methods from the TTP or cryptography fields, our method does not make use of a trust assumption or complex cryptographic protocols, but it is naturally secure due to its reliance on a federated learning algorithm which never moves the data from its original location. In the next section, we explain the concept and implementation behind our proposed privacy-preserving GLM technique.
	
	\section{Proposed method}
	\label{sec:method}
	Our proposed method uses \textit{block coordinate descent} (BCD) to estimate generalized linear models (GLM) in a situation where data is vertically partitioned across two or more parties. In BCD, parameters are iteratively updated for each block of features, cycling over the blocks until an optimum is found \citep{Hastie2015}. This optimization algorithm can be seen as a form of distributed learning \citep{bertsekas1989parallel, richtarik2016distributed} which we exploit as a privacy-preserving method because the features remain in different locations. Only linear predictions need to be transferred across the feature blocks -- the full data is never shared. 
	
	Note that for the remainder of the paper, we assume that the records of the data subjects are in the same order across databases, in line with \citet{gascon2017privacy}. Furthermore, we only consider the situation where the target attribute is available to both parties \citep{Sanil:2004:PPR:1014052.1014139}. In addition, we follow the tradition in the existing literature \citep[e.g.,][]{vaidya2008privacy, karr2010secure} to assume semi-honest adversaries: data parties will follow the protocol as described, but will still attempt to learn as much information as possible from other parties. This contrasts with malicious adversaries that can arbitrarily deviate from the protocol \citep{lindell2005secure}.
	
	In this section, we build up the BCD algorithm from the simpler case of linear regression before extending it to full GLM. Therefore, we first explain the necessary background on linear regression, as well as the notation used throughout this paper. Then, coordinate descent estimation is introduced as a means to estimate its maximum likelihood coefficients. In Section \ref{sec:bcd}, this algorithm is then extended to accommodate a vertically partitioned data structure, and in Section \ref{sec:log} we generalize it to different outcome families in order to estimate GLMs. Finally, we develop a novel method to obtain standard errors within this framework.
	
	\subsection{Background}
	We consider the centered design matrix with features $\boldsymbol{X} \in \mathbb{R}^{N\times P}$ and the centered target variable $\boldsymbol{y} \in \mathbb{R}^{N \times 1}$, where $N$ is the sample size, or number of observations, and $P$ is the number of features. The $p^{th}$ column in $\boldsymbol{X}$ is represented as $\boldsymbol{x}_p$. The columns in $\boldsymbol{X}$ excluding the $p^{th}$ are denoted as $\boldsymbol{X}_{\text{-} p}$.
	
	The basic regression model is then as follows:
	\begin{equation}
	\label{eq:regr}
	\boldsymbol{y} = \boldsymbol{X\beta} + \boldsymbol{\epsilon}
	\end{equation}
	where $\boldsymbol{\beta} \in \mathbb{R}^{P}$, $\boldsymbol{\epsilon} \sim \mathcal{N}(0, \sigma^2\boldsymbol{I})$, and $\boldsymbol{\epsilon} \perp \boldsymbol{X}$. The well-known closed-form maximum likelihood estimator of the $P$ regression coefficients $\boldsymbol{\beta}$ in this model is:
	\begin{equation}
	\label{eq:bhat}
	\boldsymbol{\hat{\beta}} = (\boldsymbol{X}^T\boldsymbol{X})^{-1}\boldsymbol{X}^T\boldsymbol{y}
	\end{equation}
	We further define the vector of predicted values as $\boldsymbol{\hat{y}} = \boldsymbol{X\hat{\beta}}$ and the vector of residuals as $\boldsymbol{\hat{\epsilon}} = \boldsymbol{y} - \boldsymbol{\hat{y}}$. 
	
	\subsection{Cyclic coordinate descent estimation}
	When instead of the full design matrix $\boldsymbol{X}$ we consider only the $p^{th}$ variable, the estimator in Equation \ref{eq:regr} yields the \textit{marginal} regression coefficient. Thus, by simplifying Equation \ref{eq:regr} to the univariate case, the marginal coefficient for the $p^{th}$ variable $\beta^*_p$ is estimated as
	\begin{equation}
	\label{eq:marg}
	\hat{\beta}^*_p = \frac{\langle \boldsymbol{x}_p, \boldsymbol{y} \rangle}{\langle \boldsymbol{x}_p\,, \boldsymbol{x}_p \rangle} = \frac{\mathrm{cov}(\boldsymbol{x}_p\,, \boldsymbol{y})}{\mathrm{var}(\boldsymbol{x}_p)}
	\end{equation}
	where $\langle \cdot\,, \cdot \rangle$ indicates the inner product of two vectors. The covariance/variance notation holds because we assume a centered design matrix $\boldsymbol{X}$ and outcome variable $\boldsymbol{y}$. 
	
	If $\boldsymbol{x}_p$ covaries with any of the predictors in $\boldsymbol{X}_{\text{-} p}$, the marginal coefficient $\boldsymbol{\beta}^*_p$ is different from the \emph{conditional} coefficient $\boldsymbol{\beta}_p$. The estimate of this coefficient is an element of $\boldsymbol{\hat{\beta}}$ in Equation \ref{eq:regr}, but it can equivalently be estimated in a coordinate-wise, univariate manner \citep{Hastie2015} as follows:
	\begin{equation}
	\label{eq:coord}
	\hat{\beta}_p = \frac{\langle \boldsymbol{x}_p, \boldsymbol{\boldsymbol{\hat{\epsilon}}_{\text{-} p}} \rangle}{\langle \boldsymbol{x}_p\,, \boldsymbol{x}_p \rangle} = \frac{\langle \boldsymbol{x}_p\,, \boldsymbol{y} - \boldsymbol{X}_{\text{-} p}\boldsymbol{\hat{\beta}}_{\text{-} p} \rangle}{\langle \boldsymbol{x}_p\,, \boldsymbol{x}_p \rangle} =  \frac{\langle \boldsymbol{x}_p\,, \boldsymbol{y}\rangle}{\langle \boldsymbol{x}_p\,, \boldsymbol{x}_p \rangle} - \frac{\langle \boldsymbol{x}_p\,,  \boldsymbol{X}_{\text{-} p}\boldsymbol{\hat{\beta}}_{\text{-} p} \rangle}{\langle \boldsymbol{x}_p\,, \boldsymbol{x}_p \rangle}
	\end{equation}
	
	The residual $\boldsymbol{\hat{\epsilon}}_{\text{-} p} = \boldsymbol{y} - \boldsymbol{X}_{\text{-} p}\boldsymbol{\hat{\beta}}_{\text{-} p}$ is the residual with respect to the variables excluding $\boldsymbol{x}_p$, evaluated at the maximum likelihood (ML) estimates of $\boldsymbol{\beta}$. Equation \ref{eq:coord} states that the conditional regression coefficient can be obtained by computing the marginal regression coefficient of $\boldsymbol{\hat{\epsilon}}_{\text{-} p}$ on $\boldsymbol{x}_p$. This relation holds because $\boldsymbol{\hat{\epsilon}}_{\text{-} p}$ represents the part of the outcome variable unrelated to $\boldsymbol{X}_{\text{-} p}$ -- by definition, $\boldsymbol{\hat{\epsilon}}_{\text{-} p} \perp \boldsymbol{X}_{\text{-} p}$. In addition, the last part of Equation \ref{eq:coord} shows that the marginal and conditional estimate of the $p^{th}$ regression coefficient are equal if $\boldsymbol{x}_p$ and $\boldsymbol{X}_{\text{-} p}$ do not covary, because the last term drops out.
	
	%
	
	The coordinate-wise estimation of $\boldsymbol{\hat{\beta}}_p$ (Equation \ref{eq:coord}) requires the maximum likelihood estimates  $\boldsymbol{\hat{\beta}}_{\text{-} p}$ of the remaining variables to be known. However, when estimation of $\boldsymbol{\hat{\beta}}$ is the goal, these estimates are not available. This can be solved by an iterative updating procedure of the $\boldsymbol{\hat{\beta}}$ estimates:
	
	\begin{minipage}{0.8\linewidth}
		\vspace{0.5cm}
		\textbf{Algorithm 1: Cyclic coordinate descent}
		\label{alg:ccd}
		
		\citep{Hastie2015}
		\begin{enumerate}
			\item Initialize $\boldsymbol{\hat{\beta}} \gets \boldsymbol{\hat{\beta}}^*$ (marginal coefficients)
			\item For each $p \in P$:
			\begin{enumerate}
				\item $\boldsymbol{\hat{\epsilon}}_{\text{-} p} \gets \boldsymbol{y} - \boldsymbol{X}_{\text{-} p}\boldsymbol{\hat{\beta}}_{\text{-} p}$
				\item $\hat{\beta}_p \gets \langle \boldsymbol{x}_p\,, \boldsymbol{\hat{\epsilon}}_{\text{-} p} \rangle \,/\, \langle \boldsymbol{x}_p\,, \boldsymbol{x}_p \rangle$
			\end{enumerate}
			\item Repeat step (2.) for $R$ iterations until convergence (i.e., the change in parameter estimates over iterations becomes negligible)
		\end{enumerate}
		\vspace{0.4cm}
	\end{minipage}
	
	An advantage of this method is that it does not require storing the full $P\times P$ covariance matrix in memory, and this matrix does not need to be inverted -- an $\mathcal{O}(P^3)$ operation. This advantage becomes especially relevant as $P$ grows \citep{Hastie2015}. Another advantage is that this estimation method allows for regularization to be implemented naturally. For example, the $\ell_1$ penalized parameters can be computed by soft-thresholding $\langle \boldsymbol{x}_p\,, \boldsymbol{\hat{\epsilon}}_{\text{-} p} \rangle$ in each iteration. This is the approach taken by the popular regularized regression package \texttt{glmnet} \citep{friedman2010}.
	
\begin{figure}[h]
		\centering
		\includegraphics[width=\linewidth]{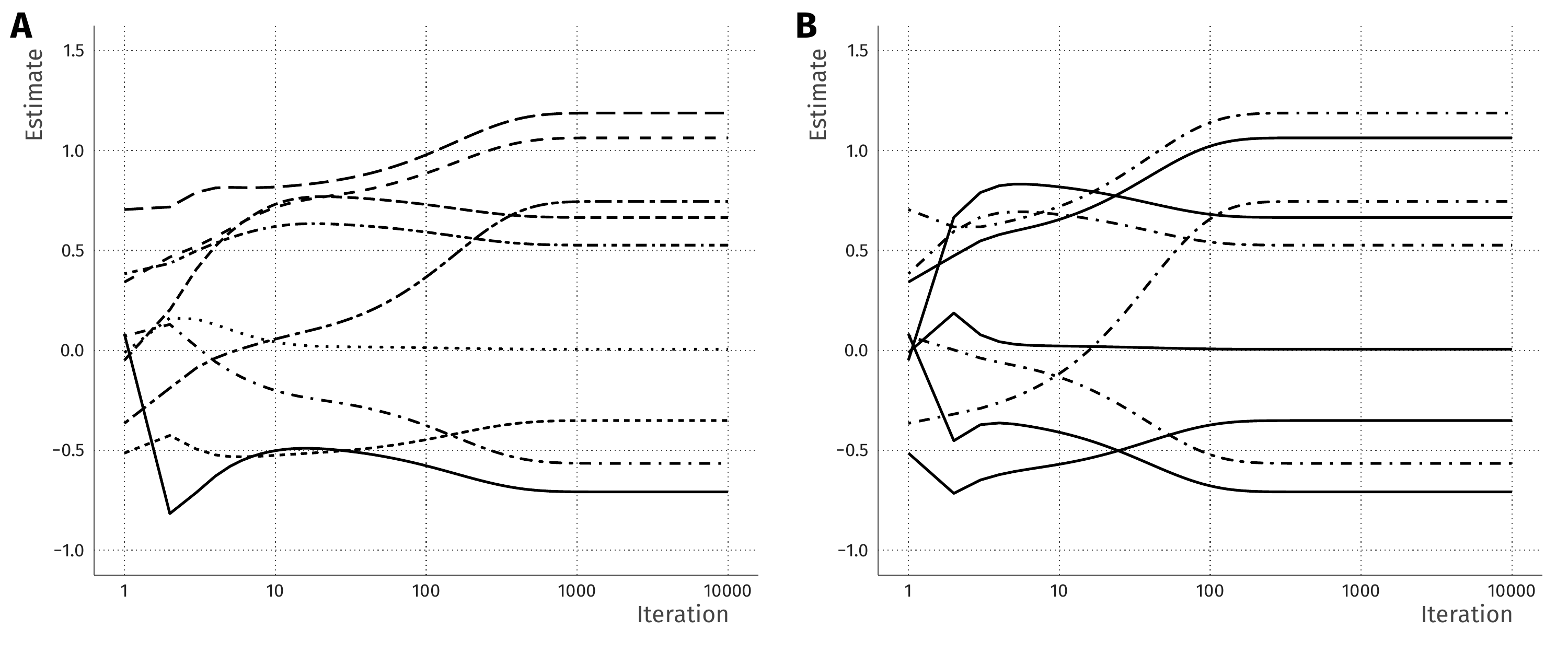}
		\caption{Panel A: Coordinate descent paths for linear regression with 9 covarying features, simulated from a multivariate normal distribution. The parameter lines converge from the marginal ML estimates (left) to the conditional ML estimates (right). Note that the x-axis is on a logarithmic scale and convergence happens around iteration 1000. Panel B: Block coordinate descent path for regression with 9 covarying predictors, applied to the same simulated dataset. There are two blocks, indicated by the line types. Note that convergence happens before iteration 500, faster than the cyclic coordinate descent algorithm.}
		\label{fig:paths}
	\end{figure}

	A graphical display of the behaviour of the estimated parameters during the cyclical coordinate descent procedure is shown in panel A of Figure \ref{fig:paths}. Here, 9 covarying features $\boldsymbol{X}$ were generated from a multivariate normal distribution. Then random parameter values $\boldsymbol{\beta}$ and random normal errors $\boldsymbol{\epsilon}$ were created and used to generate the target variable $\boldsymbol{y} = \boldsymbol{X\beta} + \boldsymbol{\epsilon}$.  
	
	Next, we show how coordinate descent generalizes to blocks of variables, and how it may be used to estimate linear regression coefficients in the vertically partitioned data scenario described above. 
	
	\subsection{Securely estimating coefficients for linear regression}
	\label{sec:bcd}
	\label{sec:secest}
	In this section, we develop the framework for analysing vertically partitioned data. Our key contribution is the combination of two observations: 
	\begin{enumerate}
		\item Coordinate descent estimation works the same for single features as well as for blocks of features -- resulting in a variant called block coordinate descent \citep[BCD;][]{Hastie2015}.
		\item Vertically partitioned data is blocked data -- the features held by \emph{Alice} can be considered the first block, and those held by \emph{Bob} the second block.
	\end{enumerate}
	
	
	
	Following these two observations, Algorithm 2 below thus provides an iterative estimator for the parameters of \emph{Alice} ($\boldsymbol{\beta}_a$) and those of \emph{Bob} ($\boldsymbol{\beta}_b$) through sharing of predictions. Predictions from \emph{Alice} are written as $\boldsymbol{\hat{y}}_a = \boldsymbol{X}_a\boldsymbol{\hat{\beta}}_a$, and the working residual with respect to \emph{Alice}, i.e., the part of $\boldsymbol{y}$ not related to the features in $\boldsymbol{X}_a$ is then $\boldsymbol{\hat{\epsilon}}_a = \boldsymbol{y} - \boldsymbol{\hat{y}}_a$.
	
	\begin{minipage}{0.8\linewidth}
		\vspace{0.5cm}
		\textbf{Algorithm 2: Secure block coordinate descent}
		\label{alg:linear}
		\begin{enumerate}
			\item Initialize $\boldsymbol{\hat{y}}_{b} \gets \boldsymbol{0}$
			\item \emph{Alice}:
			\begin{enumerate}
				\item $\boldsymbol{\hat{\epsilon}}_b \gets \boldsymbol{y} - \boldsymbol{\hat{y}}_{b}$
				\item $\boldsymbol{\hat{\beta}}_a \gets (\boldsymbol{X}_a^T\boldsymbol{X}_a)^{-1}\boldsymbol{X}^T_a\boldsymbol{\hat{\epsilon}}_b$
				\item $\boldsymbol{\hat{y}}_{a} \gets \boldsymbol{X}_a\boldsymbol{\hat{\beta}}_a$
				\item Send $\boldsymbol{\hat{y}}_{a}$ to \emph{Bob}
			\end{enumerate}
			\item \emph{Bob}:
			\begin{enumerate}
				\item $\boldsymbol{\hat{\epsilon}}_a \gets \boldsymbol{y} - \boldsymbol{\hat{y}}_{a}$
				\item $\boldsymbol{\hat{\beta}}_b \gets (\boldsymbol{X}_b^T\boldsymbol{X}_b)^{-1}\boldsymbol{X}^T_b\boldsymbol{\hat{\epsilon}}_a$
				\item $\boldsymbol{\hat{y}}_{b} \gets \boldsymbol{X}_b\boldsymbol{\hat{\beta}}_b$
				\item Send $\boldsymbol{\hat{y}}_{b}$ to \emph{Alice}
			\end{enumerate}
			\item Repeat step (2.) and (3.) for $R$ iterations until convergence.
		\end{enumerate}
		\vspace{0.4cm}
	\end{minipage}
	
	Upon convergence, the concatenated parameter estimates vector $(\boldsymbol{\hat{\beta}}_a, \boldsymbol{\hat{\beta}}_b)$ is equal (up to a small predetermined tolerance value) to the parameter estimates vector that would be obtained using the standard maximum likelihood estimator in the combined data set \citep{tseng2001convergence}. It follows that the element-wise summed prediction $\boldsymbol{\hat{y}}_{a} + \boldsymbol{\hat{y}}_{b}$ is equal to the prediction $\boldsymbol{\hat{y}}$ that would be obtained from the combined dataset. Thus, prediction can be done without sharing the parameter estimates. Further analysis of the privacy-preserving properties of this procedure is discussed in Section \ref{sec:priv}.
	
	In panel B of Figure \ref{fig:paths} we illustrate BCD, applied to the same data set as in panel A. However, instead of $P$ blocks of 1 feature each, now there are two blocks with 5 and 4 features. BCD reaches convergence with fewer iterations than the cyclic version, because it uses more information about the covariance between the features. In general, convergence is obtained faster with fewer blocks, and with less covariance between blocks \citep{richtarik2016distributed}. In the case of orthogonal blocks, only a single iteration is needed for convergence as the marginal estimates equal the conditional estimates. \citet[][Theorem 8]{li2017faster} derived a general result about the iteration complexity of BCD, showing that for smooth convex losses such as the GLM log-likelihood, the number of iterations required for convergence is linear in the number of features $P$.\\
	
	
	In the next section, we show how our BCD approach may be modified to estimate generalized linear model coefficients for a wide range of applications. Then, we provide a way to estimate standard errors within this framework.
	
	\subsection{Extension to generalized linear models}
	\label{sec:log}
	Extending this procedure to generalized linear models (GLM) requires a slightly different estimation approach: whereas the parameter estimates of full-data linear regression can be found analytically (Equation \ref{eq:bhat}), GLM requires an iteratively reweighted least squares (IRLS) procedure \citep{wedderburn1974quasi, green1984iteratively}. In each iteration $i$ in full-data GLM, the estimates are computed as follows:
	\begin{equation}
	\boldsymbol{\hat{\beta}}^{(i + 1)} =  (\boldsymbol{X}^T\boldsymbol{W}^{(i)}\boldsymbol{X})^{-1}\boldsymbol{X}^T\boldsymbol{W}^{(i)}\boldsymbol{z}^{(i)}
	\end{equation}
	
	Here, $\boldsymbol{W}$ is a diagonal weights matrix and $\boldsymbol{z}$ is a transformation of the target variable called the \emph{working response}, computed as
	\begin{equation}
	\boldsymbol{z}^{(i)} = \boldsymbol{\eta}^{(i)} + (\boldsymbol{y} - \boldsymbol{\mu}^{(i)})\left(\frac{d\boldsymbol{\mu}^{(i)}}{d\boldsymbol{\eta}^{(i)}}\right)
	\end{equation}
	where $\boldsymbol{\eta}^{(i)} = \boldsymbol{X\hat{\beta}}^{(i)}$ and $\boldsymbol{\mu}^{(i)}$ is a function of $\boldsymbol{\eta}^{(i)}$ as predefined in the link function \citep[e.g., logit link for logistic regression;][]{mccullagh1989generalized}. From this working response, a \emph{working residual} needs to be obtained which acts like $\boldsymbol{\hat{\epsilon}}_{\text{-} p}$ in Equation \ref{eq:coord}: a response vector orthogonal to the predictors excluding feature $p$. We define this working residual as follows \citep{friedman2010}:
	\begin{equation}
	\label{eq:workres}
	\boldsymbol{\hat{\epsilon}}_{\text{-} p} = \boldsymbol{z} - \boldsymbol{X}_{\text{-} p}\boldsymbol{\hat{\beta}}_{\text{-} p}
	\end{equation}
	Using this working residual and the usual weights matrix from GLM, the coordinate descent algorithm proceeds in a similar fashion to that of linear regression (Algorithm 1). Just as with coordinate descent for linear regression, this algorithm readily extends to a blockwise procedure, meaning it can be adapted for the private regression method as discussed in Section \ref{sec:secest}. 
	
	\subsection{Computing standard errors}
	\label{sec:ses}
	A key component of inference in regression models is obtaining a measure of sampling uncertainty about the obtained estimates, usually standard errors. Under the assumptions of maximum likelihood theory, the limiting distribution of the deviation of the parameter estimates is the following:
	\begin{equation}
	\label{eq:limit}
	\sqrt{N}(\boldsymbol{\hat{\beta}}_N - \boldsymbol{\beta}) \xrightarrow{d} \mathcal{N}(\boldsymbol{0}, \boldsymbol{\Sigma}_\beta)
	\end{equation}
	where $\boldsymbol{\Sigma}_\beta$ is the asymptotic variance-covariance matrix of $\boldsymbol{\hat{\beta}}$:
	\begin{equation}
	\label{eq:acov}
	\boldsymbol{\Sigma}_\beta = \text{var}(\boldsymbol{\hat{\beta}}) = \sigma^2(\boldsymbol{X}^T\boldsymbol{X})^{-1}\\
	\end{equation}
	In linear regression, $\hat{\sigma}^2 = \langle \boldsymbol{\hat{\epsilon}} \,, \boldsymbol{\hat{\epsilon}}\rangle/(N - P)$ and the standard errors of $\boldsymbol{\hat{\beta}}$ can be computed as
	\begin{equation}
	\label{eq:selin}
	\hat{\text{se}}_{\boldsymbol{\hat{\beta}}} = \sqrt{\text{diag}(\hat{\sigma}^2(\boldsymbol{X}^T\boldsymbol{X})^{-1})}
	\end{equation}
	
	Thus, to compute an estimate of the variance-covariance matrix of the sampling distribution of the $\boldsymbol{\hat{\beta}}$ parameters, the inverse covariance matrix of the features is needed. However, when the data is vertically partitioned, part of this covariance matrix is missing for each party. As a result, computing standard errors using the above information matrix approach is impossible for vertically partitioned data without sharing the features. \\
	
	We present a novel approach to compute standard errors of the regression coefficient through creating a substitute $\boldsymbol{V}_b$ of the partner's data matrix $\boldsymbol{X}_b$. This substitute is then used as the partner's data in the computation of the asymptotic variance-covariance matrix as in Equation \ref{eq:acov}. 
	
	The substitute $\boldsymbol{V}_b$ needs to contain the same information for the parameters of Alice as the real data. This information is in the predictions received from Bob -- the parameter estimates of Alice depend only on Bob's linear predictions. Consider the inputs and outputs of Bob, as seen by Alice: as the coordinate descent algorithm progresses along the $R$ iterations, Alice can create two $N \times R$ matrices, $\boldsymbol{\hat{E}}_a$ and $\boldsymbol{\hat{Y}}_{b}$
	
	\begin{equation}
	\begin{split}
	\boldsymbol{\hat{E}}_a &= \left[\boldsymbol{\hat{\epsilon}}^{(1)}_a,\, \ldots,\, \boldsymbol{\hat{\epsilon}}^{(R)}_a\right]\\
	\boldsymbol{\hat{Y}}_{b} &= \left[\boldsymbol{\hat{y}}^{(1)}_{b},\, \ldots,\, \boldsymbol{\hat{y}}^{(R)}_{b}\right]
	\end{split}
	\end{equation}
	
	These are the input and output matrices, respectively, from the projection that Bob applies in each iteration. This projection is commonly known as the \textit{hat matrix} $\boldsymbol{H}_b\in \mathbb{R}^{N\times N}$. The hat matrix relates to Bob's data matrix $\boldsymbol{X}_b$ as follows:
	\begin{equation}
	\label{eq:hatpseudo}
	\begin{split}
	\boldsymbol{\hat{Y}}_b &= \boldsymbol{H}_b\boldsymbol{\hat{E}}_a \\
	\boldsymbol{\hat{Y}}_b &= \boldsymbol{X}_b(\boldsymbol{X}_b^T\boldsymbol{X}_b)^{-1}\boldsymbol{X}_b^T\boldsymbol{\hat{E}}_a \\
	\boldsymbol{\hat{Y}}_b &= \boldsymbol{X}_b\boldsymbol{X}_b^+\boldsymbol{\hat{E}}_a
	\end{split} 
	\end{equation}
	where $\boldsymbol{X}_b^+$ indicates the Moore-Penrose generalized inverse of $\boldsymbol{X}_b$ \citep{petersen2012matrix}.
	
	Alice can compute the projection that Bob applies in each iteration $\boldsymbol{H}_b$ as follows:
	\begin{equation}
	\boldsymbol{\hat{H}}_b = \boldsymbol{\hat{Y}}_{b}\boldsymbol{\hat{E}}_{a}^{+}
	\end{equation}
	
	Across iterations, this minimum-norm solution $\boldsymbol{\hat{H}}_b$ performs the same projection as the true hat matrix of Bob. Using this projection, Alice can then create the data substitute $\boldsymbol{V}_b \in \mathbb{R}^{N\times P_b}$. For this, $\boldsymbol{V}_b$ should have the property $\boldsymbol{\hat{H}}_b = \boldsymbol{V}_b\boldsymbol{V}_b^+$. Such a $\boldsymbol{V}_b$ has the same effect on the coefficient estimates of Alice that $\boldsymbol{X}_b$ does, because it generates the same predictions that Bob does:
	\begin{equation}
	\label{eq:decomp}
	\begin{split}
	\boldsymbol{\hat{Y}}_b &= \boldsymbol{\hat{H}}_b\boldsymbol{\hat{E}}_a \\
	\boldsymbol{\hat{Y}}_b &= \boldsymbol{V}_b\boldsymbol{V}_b^+\boldsymbol{\hat{E}}_a
	\end{split} 
	\end{equation}
	There is no unique solution to decomposing $\boldsymbol{\hat{H}}_b$ into an $N\times P$ matrix $\boldsymbol{V}_b$ and its pseudoinverse. However, a numerically convenient $\boldsymbol{V}_b$ solution can be found as the first $P_b$ eigenvectors of $\boldsymbol{\hat{H}}_b$. This is a convenient choice, because the columns of $\boldsymbol{V}_b$ are then orthogonal, meaning they also have the following property: $\boldsymbol{V}_b^+ = (\boldsymbol{V}_b^T\boldsymbol{V}_b)^{-1}\boldsymbol{V}_b^T = \boldsymbol{I}^{-1}\boldsymbol{V}_b^T = \boldsymbol{V}_b^T$. As follows from Equations \ref{eq:hatpseudo} and \ref{eq:decomp}, the $\boldsymbol{V}_b$ matrix relates to $\boldsymbol{X}_b$ by means of an unknown positive definite rotation matrix $\boldsymbol{V}_b = \boldsymbol{R}\boldsymbol{X}_b$ \citep{Pavel2019}.
	
	By leveraging this similarity of $\boldsymbol{V}_b$ to $\boldsymbol{X}_b$, Alice can create an augmented data matrix of the following form: $\boldsymbol{Z}_a = \left[ \boldsymbol{X}_a, \boldsymbol{V}_b \right]$. The augmented data matrix replaces the full data matrix in the computation of the asymptotic covariance matrix: $\boldsymbol{\Sigma}^{(a)}_\beta = \sigma^2(\boldsymbol{Z}_a^T\boldsymbol{Z}_a)^{-1}$. The partition of $\boldsymbol{\Sigma}^{(a)}_\beta$ belonging to $\boldsymbol{\beta}_a$ is then identical to its counterpart from the full data asymptotic covariance matrix $\boldsymbol{\Sigma}_\beta$ (for proof see Appendix \ref{app:proof}). The square root of its diagonal elements are thus the correct standard errors that would be obtained had the full data been available.\\
	
	Alternative standard error procedures are available, e.g., profile likelihood methods or bootstrapping, but those require additional iterations of the main block coordinate descent algorithm. This yields additional information leakage and dramatically increases time requirements. Conversely, in the novel procedure we suggest here, both parties efficiently leverage the information in the existing iterations to compute standard errors without additional communication.
	
	
	\section{Privacy analysis for block coordinate descent}
	\label{sec:priv}
	
	In this section, we analyze the information transfer within our protocol for privacy-preserving regression based on block coordinate descent. In line with previous work on this topic \citep[e.g.][]{gambs2007privacy, gascon2017privacy, vaidya2003privacy, vaidya2005privacy, vaidya2008privacy}, we take the viewpoint of semi-honest parties: \emph{Alice} and \emph{Bob} follow the protocol accurately, though they may be curious and aim to recover the other party's data. In this section, we aim to identify how well \emph{Bob} can approximate \emph{Alice}'s data using a \textit{model inversion attack} \citep{fredrikson2015model, wang2015regression}.\\
	
	\subsection{Information transfer in vertically partitioned regression}
	Information about features cannot only leak through dataset sharing, but also via sharing statistics based on this data. For example, a simple method for regression without explicitly sharing the full dataset is that by \citet{karr2009privacy}, who compute the covariance matrix of $\boldsymbol{X}$ using secure inner-product methods and share it between \emph{Alice} and \emph{Bob}. This covariance matrix allows even a semi-honest \emph{Alice} to (a) know how many features are used by \emph{Bob} and -- in the case of categorical predictors -- know how many categories there are, (b) predict the values of the features held by \emph{Bob} based on the values of the features held by \emph{Alice}, (c) compute standard errors around this prediction, and (d) compute an $R^2$ value for this prediction. In other words, in a shared covariance matrix setting \emph{Alice} can know up to a certain degree the values on each of \emph{Bob}'s features for each row in the dataset, and \emph{Alice} can know how good this prediction is. Moreover, each additional feature entered by \emph{Alice} improves the prediction of features at \emph{Bob} by definition.
	
	Thus, sharing the full covariance matrix is undesirable for privacy-preserving regression. Newer methods \citep[e.g.,][]{du2004privacy, gascon2017privacy} result in additive shares of $\text{cov}(\boldsymbol{X})$ at \emph{Alice} and \emph{Bob}, without either of them possessing the full covariance matrix. Afterwards, separate secure multiparty matrix inversion protocols or linear system solvers are used to compute the regression parameters according to Equation \ref{eq:bhat}. This generally requires complex protocols involving multiple parties, but has been argued to be a secure procedure for obtaining parameter estimates for linear regression with vertically partitioned data. In these protocols, it is clear that information transfer does occur (because the full-data estimates are obtained) but its extent is not made explicit: it is unclear how the additive shares of the covariance matrix (the ``statistics") relate to the collaborator's data -- and thus it is unclear whether that data can be reconstructed.
	
	Conversely, in our protocol the covariance matrix of the combined data is never explicitly computed. Our method uses a different ``statistic": predictions $\hat{\boldsymbol{y}}$ over $R$ iterations. Each of the $R$ predictions are computed as follows by \emph{Alice}:
	\begin{equation}
	\boldsymbol{\hat{y}}^{(r)}_{a} = \boldsymbol{X}_a\hat{\boldsymbol{\beta}}^{(r)}_a
	\label{eq:preds}
	\end{equation}
	This prediction vector is then sent to \emph{Bob}: the main information transfer. In this protocol, how this information transfer relates to \emph{Alice}'s data is thus explicit. As a result, clear conclusions can be made as to the potential for data recovery.
	
	In the case where \emph{Alice} enters only a single continuous feature in the analysis protocol, the information contained in $\boldsymbol{\hat{y}}_{a}$ is sufficient for \emph{Bob} to reproduce the values of this feature up to a multiplicative constant: $\boldsymbol{\hat{y}}_{a} = \boldsymbol{x}_a \cdot \hat{\beta}_a$. With more than one feature per party, $\boldsymbol{\hat{\beta}}_a$ becomes a vector, meaning the problem of recovering the values of any feature at \emph{Alice} is underidentified. Moreover, if the protocol is followed precisely, \emph{Bob} does not know the number of features $P$ entered into the model, meaning there is additional uncertainty about the values of $\boldsymbol{X}_a$ on the part of \emph{Bob}. In its most basic form, the protocol is therefore fully secure for semi-honest parties against reconstruction of the privacy-sensitive data matrices.
	
	\subsection{Data reconstruction using shared metadata}
	In practice, there are many situations where the basic algorithm does not suffice and metadata about $\boldsymbol{X}_a$ should be shared with \emph{Bob}. For example, to circumvent multicollinearity and non-convergence, none of the features entered into the model by \emph{Alice} should be entered by \emph{Bob}. Moreover, when distributing the model results is a goal of the analysis, it is relevant to investigate how sharing parameter estimates in addition to the predictions that are already shared leads to information transfer about the original data.
	
	In our protocol, \emph{Alice} sends $R$ predictions to \emph{Bob}. These individual predictions can be appended in a columnwise fashion to create an $N \times R$ matrix $\boldsymbol{\hat{Y}}_{a} = [\boldsymbol{\hat{y}}^{(1)}_{a},\, \ldots,\, \boldsymbol{\hat{y}}^{(R)}_{a}]$. Each prediction has an associated set of parameter estimates known only by \emph{Alice} $\boldsymbol{\hat{\beta}}^{(r)}_a$, which can be combined in a similar way to create the matrix $\hat{\boldsymbol{B}}_a \in \mathbb{R}^{P\times R}$. These relate to the data matrix at \emph{Alice} as follows:
	
	\begin{equation}
	\boldsymbol{\hat{Y}}_{a} = \boldsymbol{X}_a\hat{\boldsymbol{B}}_a
	\label{eq:xapprox}
	\end{equation}
	
	In our protocol, all of $\boldsymbol{\hat{Y}}_{a}$ is shared with \emph{Bob}, and only the $R^{th}$ column of $\hat{\boldsymbol{B}}_a$ -- the final model result -- is shared. Using these estimates, \emph{Bob} can make a rank-1 minimum-norm approximation of the data held by \emph{Alice}:
	\begin{equation}
	\boldsymbol{\hat{X}}^{(1)}_a = \boldsymbol{\hat{y}}^{(R)}_{a}\hat{\boldsymbol{\beta}}^{(R)+}_a
	\end{equation}
	where $^+$ indicates the Moore-Penrose inverse \citep{petersen2012matrix}. We show empirically in Appendix \ref{app:MSE} that using this method with one set of shared parameter estimates reveals a proportion $1/P_a$ of the variance in the data to \emph{Bob}. Only the combination of predictions and their associated parameter values allows (partial) model inversion and reconstruction of the partner's data.
	
	Furthermore, as presented in Section \ref{sec:ses}, the predictions sent to and received from \emph{Alice} can be used to create a minimum-norm approximation of the hat matrix of \emph{Alice} -- another statistic which is shared in our protocol. This hat matrix is shown in Appendix \ref{app:proof} to not contain information about the features of \emph{Alice} directly, but only about a rotation of this data such that the parameter estimates of \emph{Bob} are adequately adjusted towards the conditional estimates. 
	
	In conclusion, the protocol is secure against reconstruction of the data in the case of semi-honest parties, and sharing of the final parameter estimates $\hat{\boldsymbol{\beta}}_a$ reveals a proportion $1/P_a$ of the variance in the data to the other data party. 
	
	\subsection{Further privacy considerations}
	Purposeful attacks to recover data in the case of adversarial collaborations have not been analyzed. It is possible to design such an attack, but it is also possible to design safeguards against such attacks in the implementation of the protocol, for example based on the expected smoothness of the regression paths over iterations. We leave this analysis as a topic for further research.
	
	In addition, because of the explicit link between the shared statistics and the original data, it is possible to limit the information shared with the collaborator in several ways. For example, in each iteration \emph{Alice} may add noise to the computed parameter estimates or to the predictions sent to \emph{Bob} -- a technique from the differential privacy literature \citep{dwork2006calibrating}. Another method is to put an upper bound on the number of iterations based on the number of features in the data. This has two effects: (a) it shrinks (regularizes) the parameter estimates towards the marginal estimates and (b) it creates an upper bound $\varepsilon$ on the information shared, depending on the allowed number of iterations.\\
	
	In the next section, we show how our implementation of the BCD with vertically partitioned data performs in comparison to full-data generalized linear modeling (GLM) in simulated data as well as three real-world datasets.
	
	\section{Experiments}
	\label{sec:experiments}
	Our implementation of the BCD algorithm for vertically partitioned data is provided as an \texttt{R} package at \url{https://github.com/vankesteren/privreg}. Here, we use this implementation (version 0.9.5) to estimate models on both simulated data (Section \ref{sec:sim}) and real-world data with multiple parties from the UCI data repository (Section \ref{sec:uci}). Reproducible code for this section is available in the supplementary material to this paper.
	
	\subsection{Simulated data}
	\label{sec:sim}
	The goal of this section is to compare our proposed privacy-preserving regression method to a benchmark method under controlled conditions. The benchmark method for these experiments is linear and logistic regression with a complete dataset, since the optimum privacy-preserving method would attain the same results with vertically partitioned data. For this section, data with multiple features and one target were simulated in the R programming language \citep{RCoreTeam2018}, with the following manipulations:
	
	\begin{table}[H]
		\centering
		\normalsize
		\begin{tabular}{p{0.2\linewidth}p{0.7\linewidth}}
			\textbf{Target} & Either a normally distributed or a binomial target variable. In the case of the normal target, the $R^2$ was set to 0.5. \\
			\textbf{Dimensionality} & The total number of features was either 10, 50, or 100. \\
			\textbf{Covariance} & The covariance matrix of the features was had 1 on the diagonal and either 0.1 (low covariance) or 0.5 (high covariance) on all off-diagonal elements.
		\end{tabular}
	\end{table}
	
	For each condition, 100 datasets were randomly generated. For the privacy-preserving regression method, the generated features were then equally distributed among \emph{Alice} and \emph{Bob}, after which the estimation was started. As a baseline comparison, a generalized linear model was estimated on the full dataset with all the features using the \texttt{glm()} function from the base \texttt{R stats} package. The exact data-generating mechanism, as well as the estimation method and hyperparameters can be found in the supplementary material.
	
\begin{figure}[H]
		\centering
		\includegraphics[width=.8\linewidth]{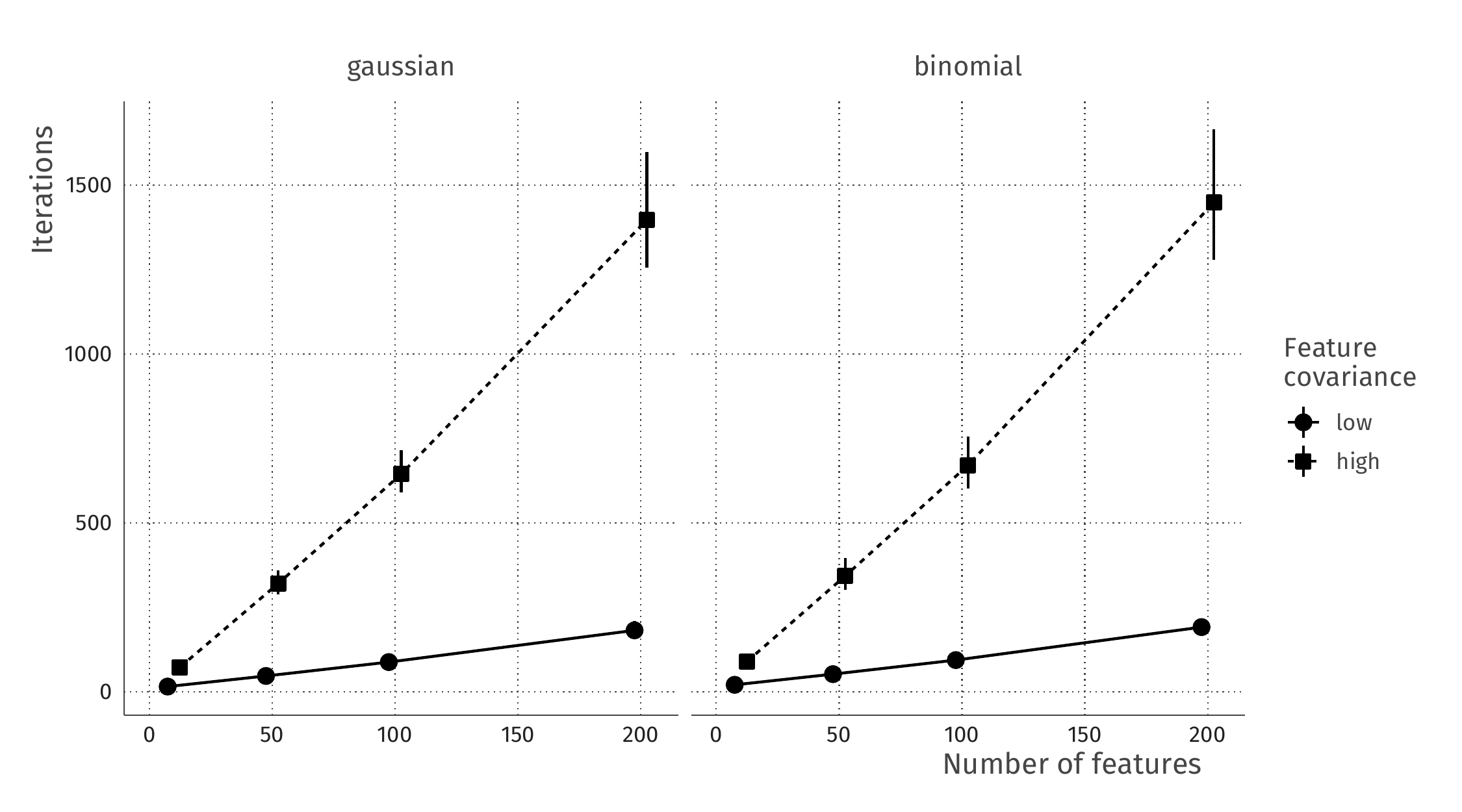}
		\caption{Observed amount of iterations required for convergence is approximately linear in the number of features and increases as there is more covariance between the features. Error bars indicate 95\% simulation percentile intervals.}
		\label{fig:iter}
	\end{figure}

	The empirical convergence rates for the privacy-preserving regression method are shown in Figure \ref{fig:iter}. As expected from the work of \citet{li2017faster}, the number of iterations required increases linearly with the number of features. In addition, the high covariance leads to slower convergence due to the conditional estimates lying further away from the marginal estimates. As mentioned in Section \ref{sec:secest}, with no covariance the number of iterations would be 1.
	
\begin{figure}[H]
		\centering
		\includegraphics[width=.8\linewidth]{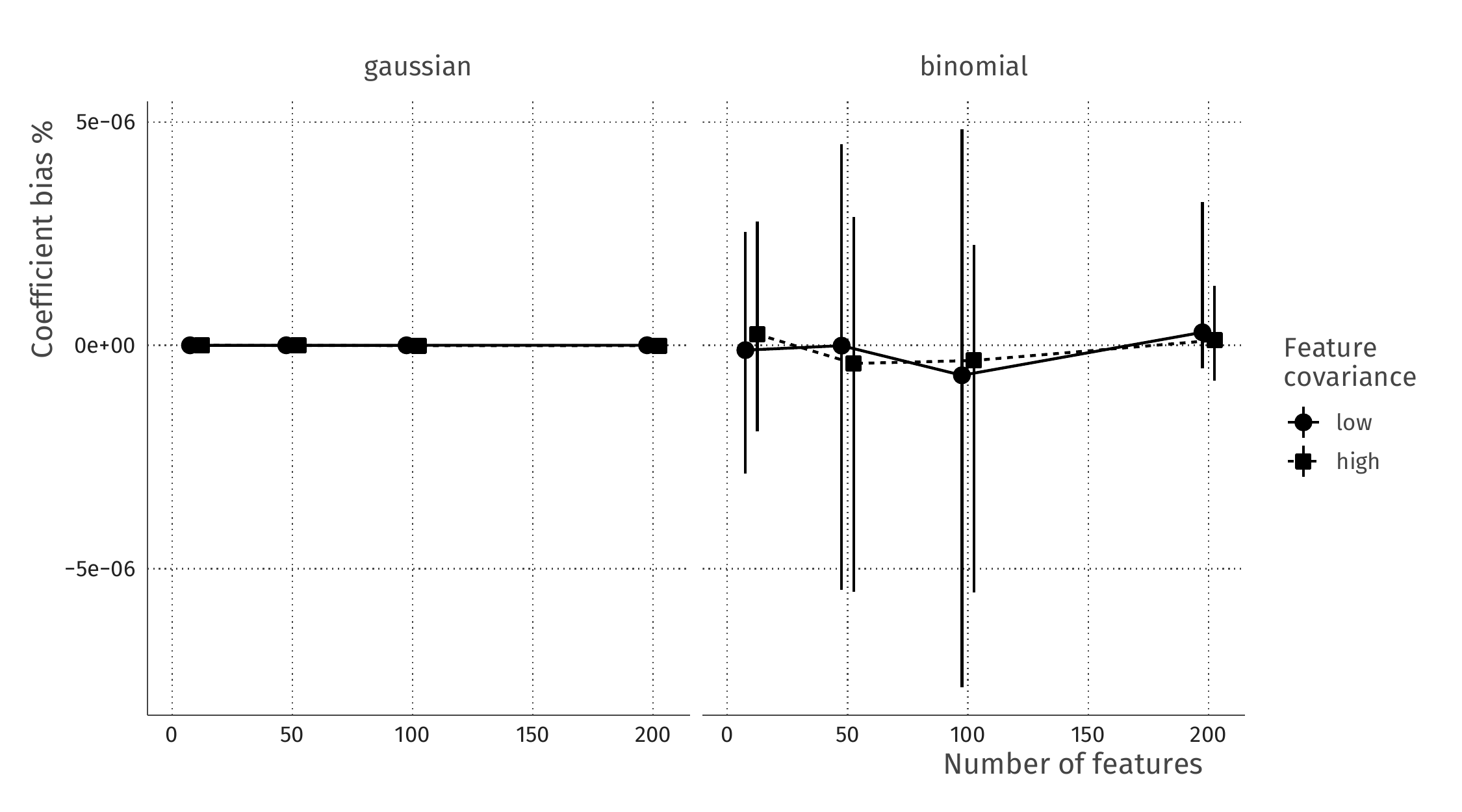}
		\caption{The parameter bias relative to the baseline GLM method is negligible for any number of features and feature covariance strength. Note the small y-axis range.}
		\label{fig:bias}
	\end{figure}

	The obtained parameter estimates ($\boldsymbol{\hat{\beta}}$) of our method are equal to those found by the baseline comparison method in all simulated conditions, up to a computational tolerance in the convergence of the estimation algorithm (Figure \ref{fig:bias}). This lack of relative bias indicates that the proposed privacy-preserving regression approach performs as well as full-data generalized linear models, at least for the extent of these simulations.
	
\begin{figure}[H]
		\centering
		\includegraphics[width=.8\linewidth]{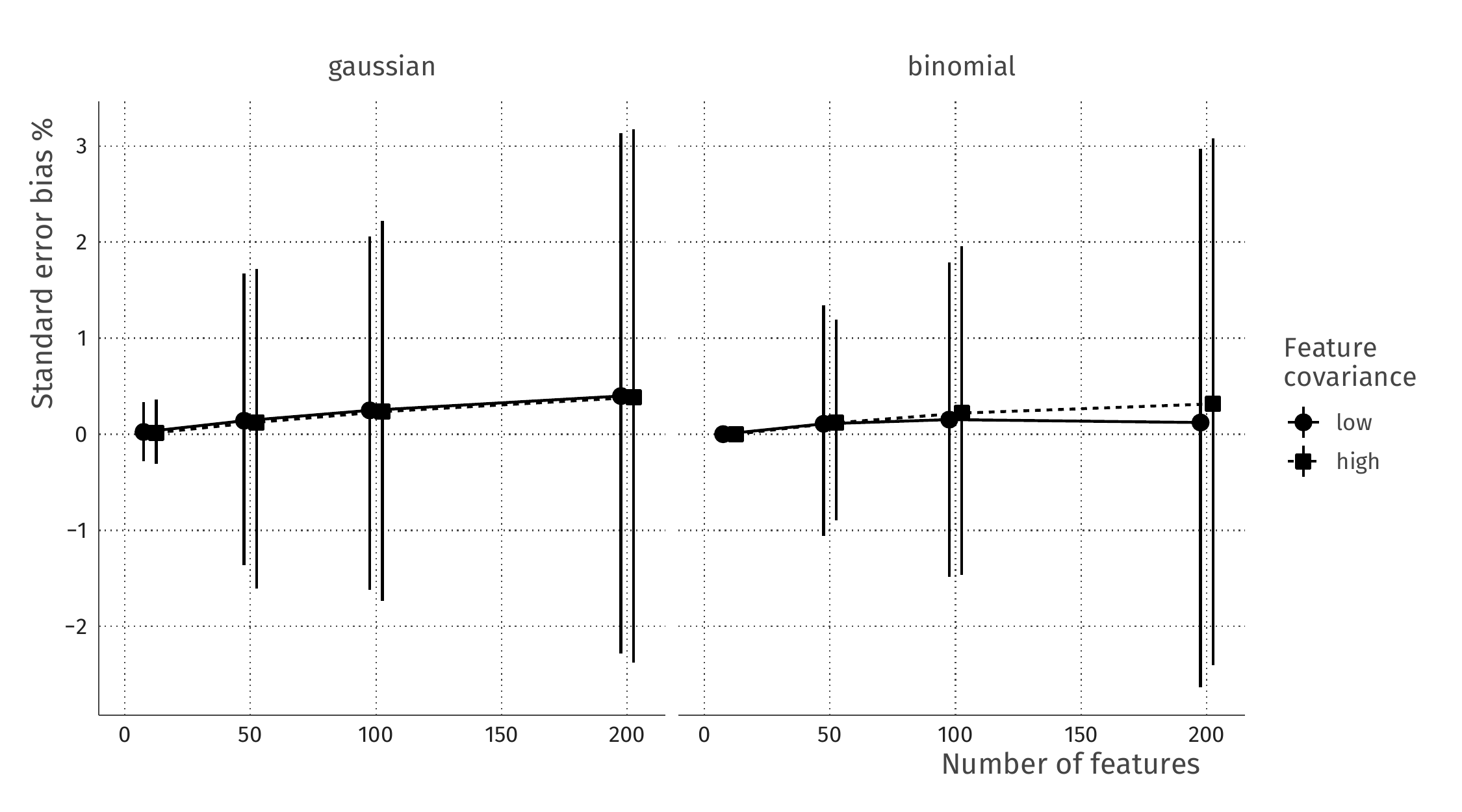}
		\caption{Standard error bias in percentage relative to the baseline GLM method. Variation across datasets increases with the number of features, and there is a very slight trend ($<1\%$) towards overestimation of the standard error for larger datasets.}
		\label{fig:se_bias}
	\end{figure}

	Standard errors indicate uncertainty in the dataset around the coefficient values, and they are the basis for statistical significance tests. Figure \ref{fig:se_bias} shows the bias in the standard errors relative to the baseline GLM method for the different conditions. The figure shows that variation of this bias over different datasets increases with the number of features (larger error bars). In addition, there seems to be a very slight relative overestimation of the standard errors on average. This is due to slightly different convergence criteria and tolerances for both methods, which propagates through the standard error procedure (Section \ref{sec:ses}). Despite this, the standard error bias is overall small ($<3\%$).
	
	In conclusion, the simulations have shown that privacy-preserving regression using block coordinate descent on vertically partitioned data has equal performance to established regression methods on full data. However, in this section the data has been simulated to behave according to specification. In the next section, we compare the performance of these two methods on real-world datasets.
	
	\subsection{UCI datasets}
	\label{sec:uci}
	In this section, we tested our proposed method on three different real-life data sets from the UCI (University of California at Irvine) Machine Learning repository \citep{blake1998uci}. The datasets were chosen because they can be naturally partitioned into two sources, and their size and targets are different (Table \ref{tab:datasets}). As before, the full preprocessing and analysis code for this section is available in the supplementary materials. Analyses were run on two separate computers (an Intel Core i7-8750H at 2.20 GHz and an Intel Xeon E5-2650 v4 at 2.20GHz) connected via a gigabit Ethernet connection on a university network.
	
	\begin{table}[H]
		\centering
		\begin{tabular}{  l  c  c  c  l } 
			\hline
			Dataset & Features & Instances & Task & Parties\\ 
			\hline 
			Forest fire & 13 & 517 & Regression & Weather \& Fire dept.\\ 
			HCC & 49 & 165 & Classification & Lab \& Clinic\\ 
			Diabetes & 43 & 15 000 & Classification & Clinic \& Pharmacy \\ 
			\hline
		\end{tabular}
		\caption{Properties of the datasets used from the UCI machine learning repository after dataset cleaning and pre-processing. Code can be found in the supplementary materials.}
		\label{tab:datasets}
	\end{table}
	
	\subsubsection{Forest fires data}
	The forest fire data comes from the Montesinho natural park in Portugal \citep{cortez2007data}. It contains several weather observations by a meteorological station (e.g. wind speed, temperature, relative humidity, etc) as well as fire department risk assessments. In this dataset, the target is to predict the area of forest burned by a particular fire using the features from the aforementioned parties. 
	
\begin{figure}[H]
		\centering
		\includegraphics[width=\linewidth]{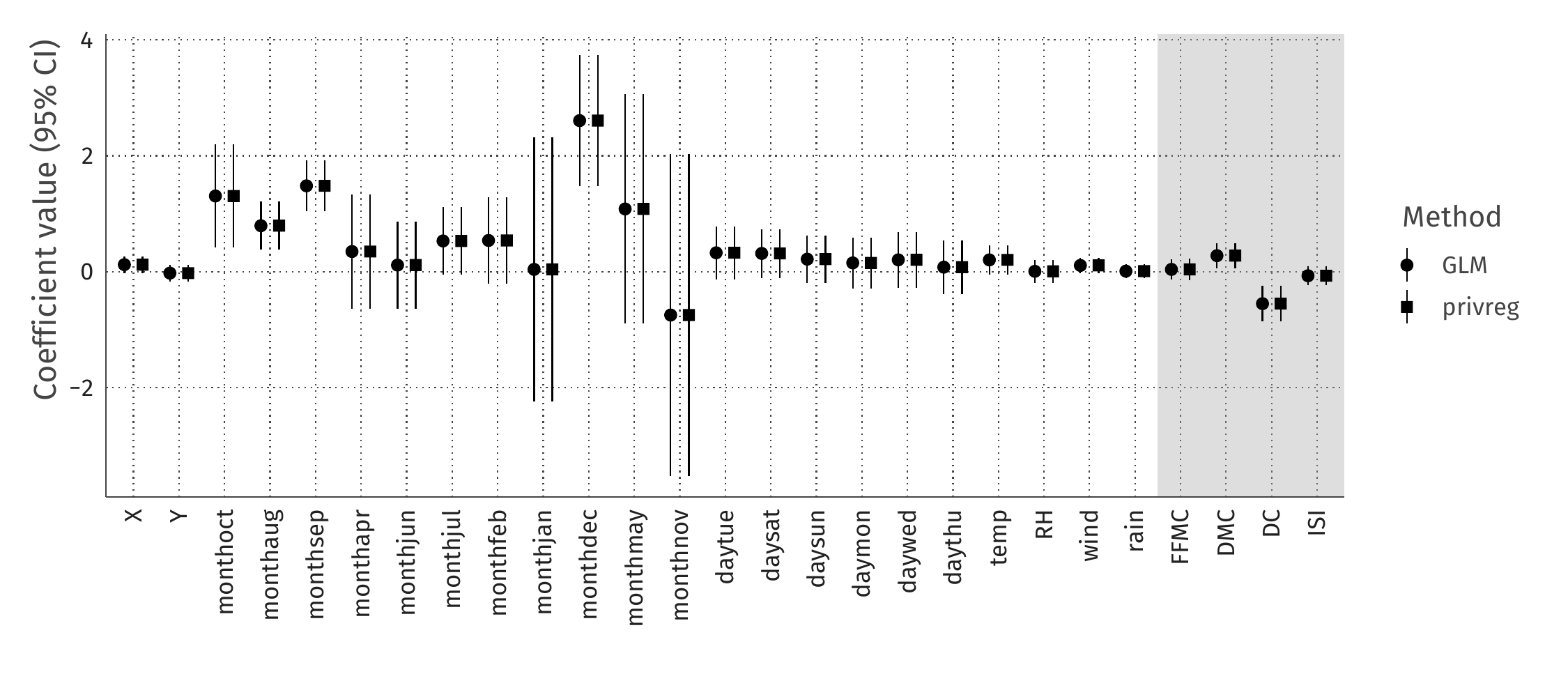}
		\caption{The coefficients and standard errors for the forest fire analysis are exactly the same for the GLM and our privacy-preserving regression estimation methods. The shading indicates data partitioning into the weather service (light) and fire department (dark).}
		\label{fig:forest}
	\end{figure}

	We performed linear regression where the target was log-transformed to normalize the residuals. Continuous features were standardized before they were entered into the analysis. The analysis took 450 BCD iterations in the privacy-preserving regression case. Including encryption and networking overhead, estimation took 14.51 seconds and computing standard errors took 0.61 seconds. Figure \ref{fig:forest} shows that the coefficients and their 95\% confidence intervals are equal for the full-data analysis and the privacy-preserving procedure. Several months show a significant positive effect on the log-area, meaning that -- conditional on the ratings of the fire department -- fires in these months (e.g., August and December) burn larger areas of forest.
	
	\subsubsection{Hepatocellular carcinoma data}
	This dataset was collected by Coimbra’s Hospital and University Centre in Portugal for studying an epithelial cell cancer of the liver called hepatocellular carcinoma (HCC) \citep{santos2015new}. It contains heterogeneous data on demographics, risk factors, laboratory and overall survival features from HCC patients. The goal of the analysis is to use lab results for a tissue sample as well as clinical data for the patient to predict survival after diagnosis. Since survival is a binary target, a binomial family GLM (logistic regression) was performed. For this analysis, continuous features were standardized before the analysis, which improved the convergence characteristics. The privacy-preserving GLM converged in 1636 iterations. Including encryption and networking overhead, estimation took 3 minutes and 16 seconds and computing standard errors took 0.63 seconds.
	
\begin{figure}[H]
		\centering
		\includegraphics[width=\linewidth]{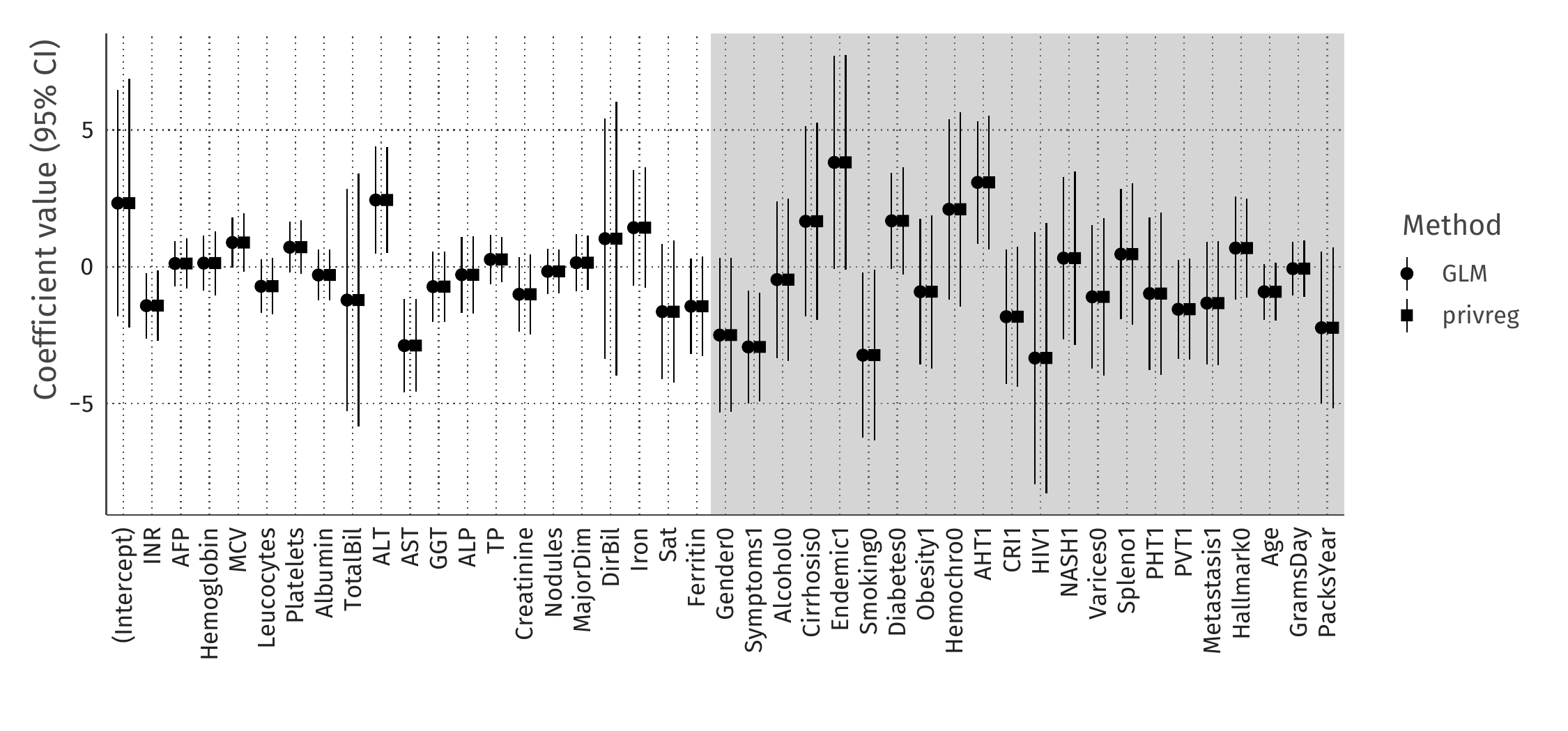}
		\caption{The coefficients and standard errors for the carcinoma analysis are very similar for the GLM and our privacy-preserving regression estimation methods. The shading indicates data partitioning into the lab results (light) and clinic (dark).}
		\label{fig:carcinoma}
	\end{figure}

	The results of the analysis (Figure \ref{fig:carcinoma}) show that the estimates are exactly equal across the full-data and the privacy-preserving analyses, meaning survival probability predictions for new incoming patients based on these models will be the same. Despite slight deviations in the width of the confidence intervals, conclusions about the effects of the features on survival are also the same in this dataset. 
	
	\subsubsection{Diabetes}
	The diabetes dataset is an extract representing 10 years (1999-2008) of clinical diabetes care at 130 hospitals and integrated delivery networks throughout the United States \citep{strack2014impact}. It is a large and also heterogeneous data set including encounter data (emergency, outpatient, and inpatient), provider speciality, demographics, laboratory data, pharmacy data, in-hospital mortality, and hospital characteristics. In this dataset, we predict readmission to the hospital using both administrative features and pharmaceutical features. To keep the computation of the standard errors for this analysis possible, 15000 patients were randomly selected from the dataset. Features were re-coded where necessary, and categorical features with only a single category in the sample were excluded from the analysis. The full pre-processing pipeline can be found in the supplementary material.
	
	Since readmission is a binary target, a binomial family GLM (logistic regression) was performed. The diabetes data analysis required 284 iterations of the BCD algorithm. Including encryption and networking overhead, estimation took 1 minute and 37 seconds and computing standard errors took 42 seconds. This analysis is particularly interesting with respect to the effect of insulin (\texttt{insulinYes}) on the readmission probability. In the analysis of only the medication data, insulin has a significant positive effect on readmission (OR = 1.20, $p < .001$), whereas conditional on the administrative data, insulin significantly reduces the readmission probability (OR = 0.88, $p < .001$). This is a strong argument for including the data of both parties in the analysis.
	\begin{figure}[H]
		\centering
		\includegraphics[width=\linewidth]{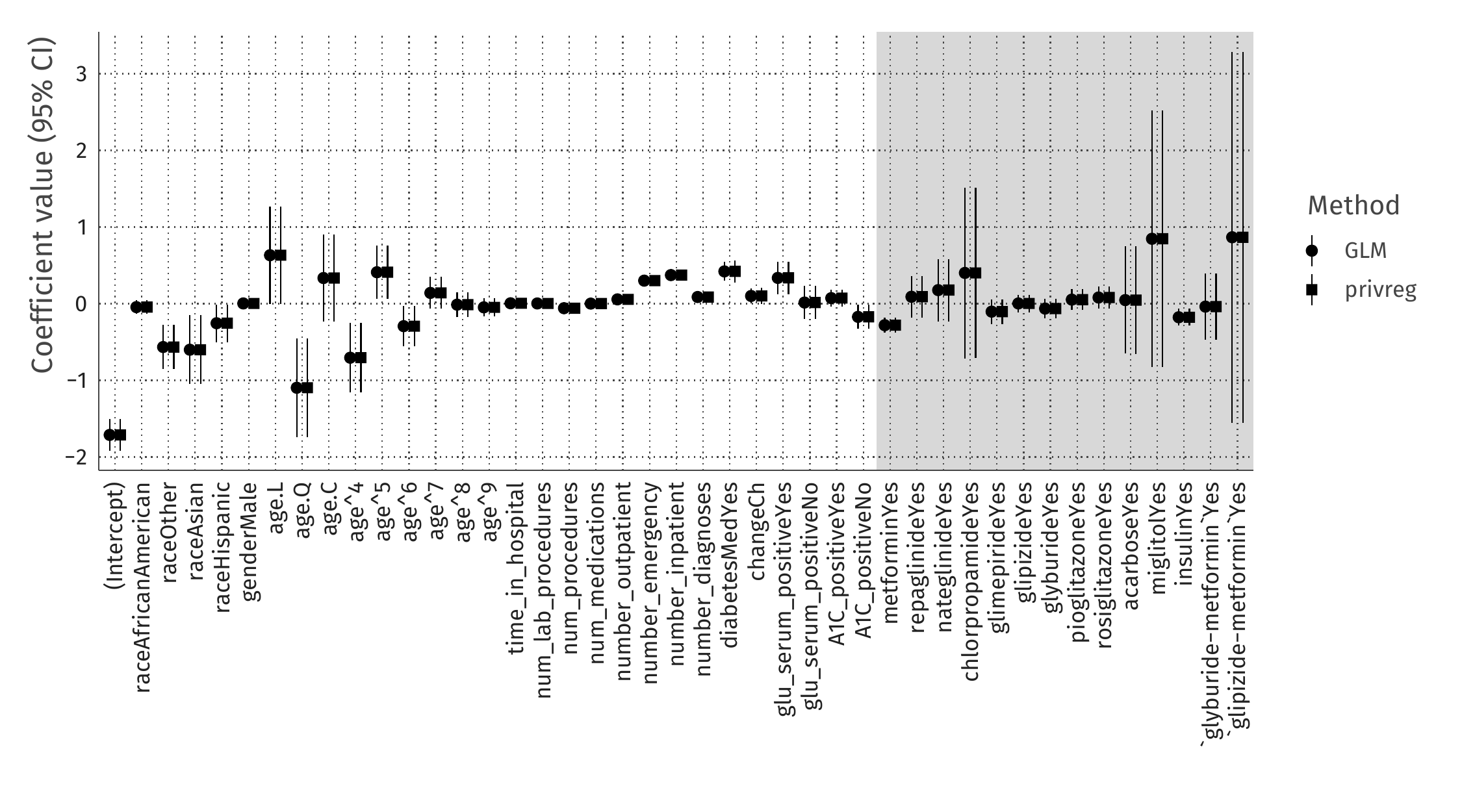}
		\caption{The coefficients and standard errors for the diabetes analysis are exactly the same for the GLM and our privacy-preserving regression estimation methods. The shading indicates data partitioning into the clinical data (light) and pharmaceutical data (dark).}
		\label{fig:diabetes}
	\end{figure}
	
	In this section, we have shown that privacy-preserving regression using block coordinate descent is not only a theoretical possibility, but also a viable implementation of GLM for analyzing data with varied characteristics -- both in simulated data under controlled conditions (Section \ref{sec:sim}) and in real-world prediction and analysis problems with various targets (Section \ref{sec:uci}). The time constraints on the real-world analyses are manageable, with all example analyses converging in under 4 minutes. We have shown that the parameter estimates exactly match those of the existing reference methods, and that our novel estimation method for the standard errors generally agrees with its full-data counterpart -- and where it did not the difference was so small that it lead to the same conclusions in the analysis.
	
	\section{Discussion}
	\label{sec:disc}
	In this paper, we have argued that block coordinate descent is a general method for estimating conditional parts of a generalized linear model (GLM) in a vertically partitioned data situation. Using this approach, two or more data parties can collaboratively estimate a GLM without sharing their features. This is useful when the features are not allowed to be shared, for example when there are privacy issues. 
	
	Our method falls within the category of federated learning algorithms. This means it can be implemented for situations when data mining is to be performed over remote devices or siloed data centers \citep{li2019federated}, where aggregating the data tables is prohibitively expensive in terms of time, computation, or storage costs. This work aligns with several recent contributions that seek to exploit the privacy-preserving aspects of federated learning algorithms \citep[see, e.g.,][]{bonawitz2016practical, geyer2017differentially}. 
	
	Due to the accessibility of our protocol and its similarity to existing regression estimation methods, extensions are relatively simple to implement. First and foremost, our framework can be extended to multiple parties as coordinate descent naturally extends to multiple blocks. In addition, our algorithm could include penalties for regularized estimation of the regression parameters through thresholding \citep{friedman2010}. Through further research into combining coordinate descent with missing data methods such as full information maximum likelihood \citep{enders2001performance}, our protocol could even be extended for a hybrid partitioning situation where data is both horizontally and vertically partitioned.
	
	Our novel approach is a natural modification of the familiar linear modeling framework -- without changes in the assumptions. We argue that our protocol restricts statistical information sharing as much as possible, while being explicit in how the shared information relates to the original data. Because of this, data parties know how much information they share, and the protocol could even incorporate methods from the differential privacy literature -- such as additive noise or early stopping -- to put a restriction on the amount of information shared with the partner institution \citep{dwork2006calibrating}.
	
	The main tradeoff of this flexibility compared to existing methods is relatively high communication cost: each iteration requires $N$ prediction values to be sent to the partner institution. In addition, like other methods for this situation the block coordinate descent assumes (probabilistic) linkage of the individual records -- both parties need to have their records in the same order. Lastly, this method is possible only when the target can be shared, although in absence of a shareable target collaborators could still perform some form of transfer learning, e.g., by predicting a shareable feature \emph{related} to the true target.
	
	Considering the prospect of these extensions and the availability of an accessible open-source implementation, we believe the proposed block coordinate descent protocol can be a springboard for future developments in the privacy-preserving data mining field.
	
	
	
	\section*{Compliance with Ethical Standards}
	\textbf{Funding}: this work was supported by the Netherlands Organization for Scientific Research (NWO) under grant number 406.17.057 and by the Dutch National Research Agenda (NWA) under project number 400.17.605.\\
	
	\noindent \textbf{Conflict of Interest}: The authors declare that they have no conflict of interest.\\
	
	\noindent \textbf{Acknowledgments}: We thank Ayoub Bagheri for his helpful comments on an earlier version of this manuscript.

	\bibliographystyle{spbasic}

	\newpage
	\appendix
	
	\section{Proof for recovery of standard errors}
	\label{app:proof}
	Let $A = X^TX$, partitioned into four submatrices $A_{11}$ (held by Alice), $A_{22}$ (held by Bob), and $A_{22}$ (unknown to either). The standard inverse of such a partitioned, positive definite symmetric matrix is
	\begin{equation}
	\begin{split}A^{-1} &= 
	\left(
	\begin{array}{cc}
	B_{11} & B_{12}\\
	B_{22} & B_{22}
	\end{array}
	\right) \\
	&=
	\left(
	\begin{array}{cc}
	\left(A_{11} - A_{12} A_{22}^{-1} A_{12}^T\right)^{-1} 
	&
	-A_{11}^{-1} A_{12} \left( A_{22} - A_{12}^T A_{11}^{-1} A_{12}  \right)^{-1}\\
	-A_{22}^{-1} A_{12}^T \left( A_{11} - A_{12} A_{22}^{-1} A_{12}^T \right)^{-1}
	&
	\left(A_{22} - A_{12}^T A_{11}^{-1} A_{12}\right)^{-1}         
	\end{array}
	\right)
	\end{split}
	\end{equation}

	Following the procedure outlined in Section \ref{sec:ses}, Alice replaces $X_{2}$ with $V_2 = R_{2} X_{2}$, and Bob replaces $X_{1}$ with $V_1 = R_{1} X_{1}$, where $R_{j}$ are unknown orthogonal rotation matrices. This gives two new matrices, $A^{(1)}$ and $A^{(2)}$, and their inverses, $B^{(1)}$ and $B^{(2)}$.
	By substition,
	\begin{equation}
	\begin{split}
	A^{(1)}_{12} &= X_1^T R_2 X_2\\
	A^{(1)}_{22} &= X_2^T R_2^T R_2 X_2
	\end{split}
	\end{equation}
	So that
	\begin{equation}
	\begin{split}
	B^{(1)}_{11}
	&= \left(A^{(1)}_{11} - A^{(1)}_{12} (A^{(1)}_{22})^{-1} (A^{(1)}_{12})^T\right)^{-1} \\
	&= \left((X_1^T X_1) - (X_1^T R_2 X_2) (X_2^T R_2^T R_2 X_2)^{-1} (X_1^T R_2 X_2)^T\right)^{-1}\\
	&= \left((X_1^T X_1) - (X_1^T X_2) (X_2^T X_2)^{-1} (X_1^T X_2)^T\right)^{-1}\\
	&= \left(A_{11} - A_{12} A_{22}^{-1} A_{12}^T\right)^{-1} \\
	&= B_{11}
	\end{split}
	\end{equation}
	This shows that the part of the usual ACOV to do with $\hat{\beta}_1$ can be estimated correctly, and therefore the standard errors are available: $\text{ACOV}(\hat{\beta_j}) = \sigma^2 B_{jj}$.
	Moreover,
	\begin{equation}
	\begin{split}
	B_{21}^{(1)} &= -(A^{(1)}_{22})^{-1} (A^{(1)}_{12})^T B_{11}\\
	&= -(R_2^T R_2)^{-1} R_2 B_{21}
	\end{split}
	\end{equation}
	so that 
	\begin{equation}
	\begin{split}
	\left[(Z^T Z)^{-1} Z^T y \right]_{p_1} &= 
	B_{11} X_1^T y -(R_2^T R_2)^{-1} R_2^T R_2 B_{21}^T X_2^T y \\
	&= B_{11} X_1^T y - B_{21}^T X_2^T y \\
	&= \hat{\beta_{1}}
	\end{split}
	\end{equation}
	This shows that the exact same estimates are obtained for $\hat{\beta_1}$. The same proof can be given for Bob and $\hat{\beta_2}$.
	
	Note further that:
	\begin{enumerate}
		\item Alice cannot get $\hat{\beta_2}$ right because $R_2$ does not drop out in the other's part of the vector
		\item We cannot get the ACOV of $(\hat{\beta_1}, \hat{\beta_2})$ for this same reason
	\end{enumerate}
	
	\newpage
	
	\section{MSE of rank-R data approximation}
	\label{app:MSE}
	From Equation \ref{eq:xapprox} we can create the following approximation:
	\begin{equation}
	\begin{split}
	\boldsymbol{\hat{Y}}_{a} &= \boldsymbol{X}_a\hat{\boldsymbol{B}}_a\\
	\boldsymbol{\hat{X}}_a &= \boldsymbol{\hat{Y}}_{a}\hat{\boldsymbol{B}}_a^+
	\end{split}
	\end{equation}
	
	where $\boldsymbol{\hat{Y}}_{a} \in \mathbb{R}^{N \times R}$, ${\boldsymbol{B}}_a \in \mathbb{R}^{P \times R}$, and $\boldsymbol{X}_a \in \mathbb{R}^{N \times P}$ and all matrices are full rank. For simplicity, but without loss of generality, we assume here that the variance of all the features in $\boldsymbol{X}_a$ is the same, $\sigma^2_a$, and these features are uncorrelated.
	
\begin{figure}[h]
		\centering
		\includegraphics[width=.66\linewidth]{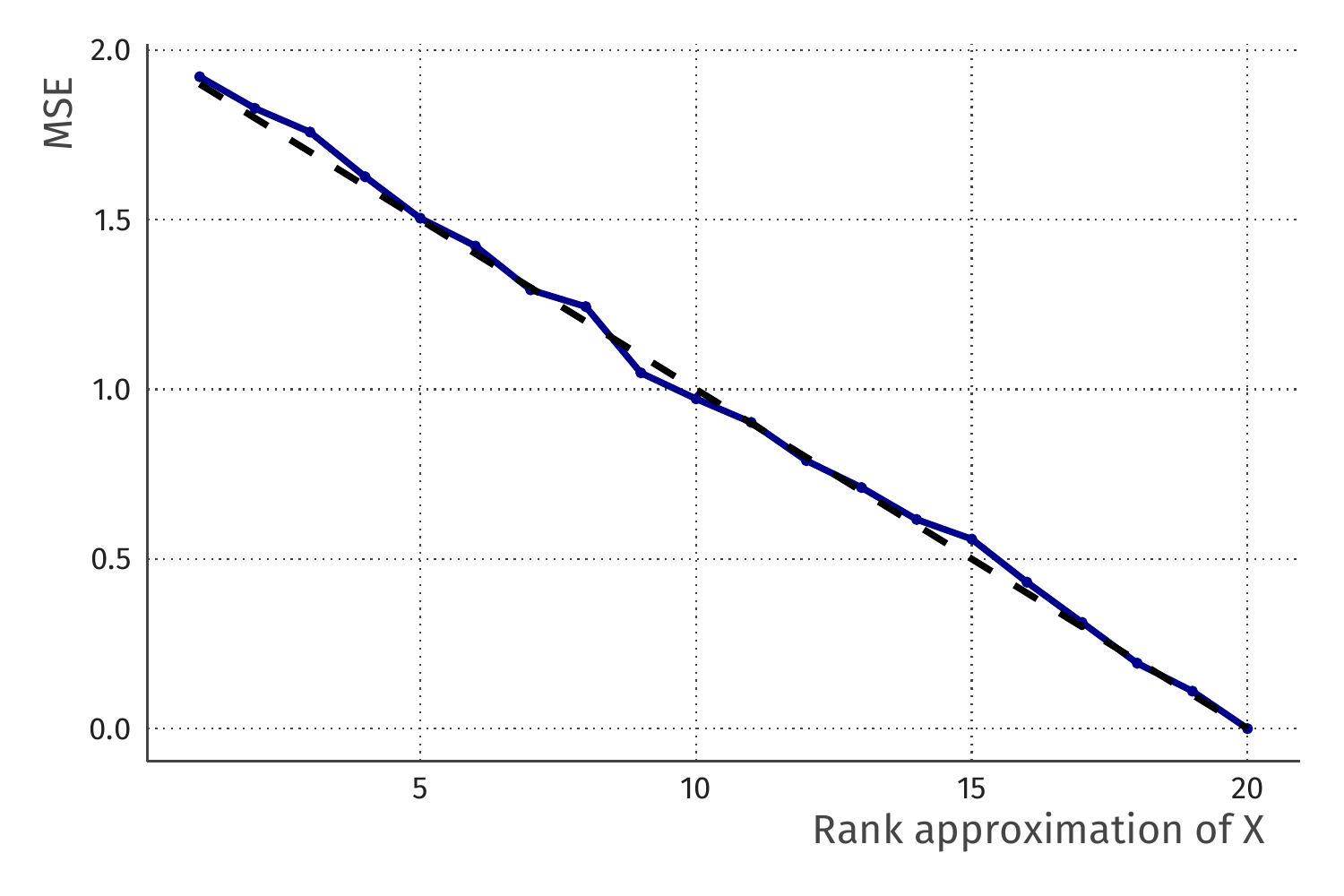}
		\caption{Mean square error (MSE) of the approximation of the data $\boldsymbol{X}_a$ at Alice by Bob if $\hat{\boldsymbol{B}}_a$ is known. $\boldsymbol{X}_a$ was simulated as having $P=20$ uncorrelated features with variance $\sigma^2_a=2$. Note that the approximation linearly improves as the rank of $\hat{\boldsymbol{B}}_a$ increases, with a perfect approximation reached when $R=P$. Dashed line indicates expected MSE, using the formula $E[\text{MSE}] = \sigma^2_a (1 - R/P)$.}
		\label{fig:rank}
	\end{figure}

	The relation between $P$, $R$, and the accuracy of the approximation $\boldsymbol{\hat{X}}_a$ is as follows: as $R \to P$, the MSE improves linearly, with perfect approximation being achieved when $R = P$. As mentioned in-text, when $P = 1$, sharing one set of parameters ($R = 1$) means the data can be recovered completely. Empirical simulations show that the relation between $R$, $P$, and expected mean square error of approximation is $\text{MSE} = \sigma^2_a (1 - R/P)$, where $\sigma^2_a$ is the variance of the features in $\boldsymbol{X}_a$ (see Figure \ref{fig:rank}).
	
	Phrasing the above in terms of information sharing and privacy preservation: in sharing $R$ sets of parameter estimates $\hat{\boldsymbol{\beta}}^{(r)}_a$ with their associated predictions $\boldsymbol{\hat{y}}^{(r)}_a$, Alice reveals a proportion of at least $R/P$ of variance in the data. This proportion is a lower bound: in case there are correlations among the features of Alice, this proportion increases. When $R = P$ the data of Alice can be reconstructed by Bob. When either of a pair $(\hat{\boldsymbol{\beta}}^{(r)}_a, \boldsymbol{\hat{y}}^{(r)}_a)$ are shared but not the other, no information is revealed.
	

\end{document}